\documentclass[journal]{IEEEtran}
\usepackage[numbers,sort&compress]{natbib}
\usepackage{amsmath}
\usepackage{amsfonts}
\usepackage{xcolor}
\usepackage{url}
\usepackage{amsmath}
\usepackage{amssymb}
\usepackage{dsfont}
\usepackage{booktabs}
\usepackage{algorithm}
\usepackage{booktabs}
\usepackage[noend]{algpseudocode}

\usepackage[pdftex]{graphicx}
\usepackage{svg}
\usepackage{soul}
\usepackage[normalem]{ulem}
\useunder{\uline}{\ul}{}

\title{Latent Space Unsupervised Semantic Segmentation}
\author{Knut J. Strommen$^{1}$, Jim~Tørresen$^{1}$, Ulysse Côté-Allard$^{1}$ %
\thanks{}
\thanks{$^{1}$Knut J. Strommen, Jim Torresen and Ulysse Côté-Allard are with the University of Oslo, Oslo, Norway. Corresponding author: Knut J. Strommen ({\tt\small knutjstr@ifi.uio.no})}%
}

\begin{document}
\maketitle
\begin{abstract}
The development of compact and energy-efficient wearable sensors has led to an increase in the availability of biosignals. 
To analyze these continuously recorded, and often multidimensional, time series at scale, being able to conduct meaningful unsupervised data segmentation is an auspicious target.
A common way to achieve this is to identify change-points within the time series as the segmentation basis. However, traditional change-point detection algorithms often come with drawbacks, limiting their real-world applicability. 
Notably, they generally rely on the complete time series to be available and thus cannot be used for real-time applications. Another common limitation is that they poorly (or cannot) handle the segmentation of multidimensional time series. 
Consequently, the main contribution of this work is to propose a novel unsupervised segmentation algorithm for multidimensional time series named Latent Space Unsupervised Semantic Segmentation (LS-USS), which was designed to work easily with both online and batch data. 
When comparing LS-USS against other state-of-the-art change-point detection algorithms on a variety of real-world datasets, in both the offline and real-time setting, LS-USS systematically achieves on par or better performances. 

\end{abstract}


\section{Introduction}

The physiological processes occurring in the human body generate a plethora of biosignals (e.g. motion, muscle activity, biopotential) that can provide otherwise inaccessible insights into a person's health and activity, and may even be leveraged for the development of human-computer interfaces~\cite{emg_control_interface}.
Through the rise of the Internet of Things and the development of more compact and energy efficient sensors, wearable technologies can now provide continuous, non-intrusive and multimodal monitoring of these biosignals in real-time. 
However, the sheer amount of data generated by these devices make the processing and analysis of this information challenging and time-consuming to perform. As an example, a single 9-axis inertial measurement unit cadenced at a low sampling rate of 60Hz generates around 2 million data-points every hour. Having to manually annotate these unlabelled data streams can thus rapidly become impractical. Another commonly occurring type of data are weakly labeled time series which include imprecise or inexact labels of when a change actually occurred. For both unlabeled and weakly labeled time series, the most useful information often lies in the precise location where a transition in the time series take place. Thus to make use of this data at scale, being able to identify these critical points in the time series in an unsupervised manner can become highly beneficial. Unsupervised change-point detection (CPD) algorithms attempt to identify these abrupt change in the data generating process~\cite{CDP_survey_2017}.

Unfortunately, CDPs algorithms also tend to suffer from limitations that reduce their suitability for real-world applications. Notably, they tend to make assumptions and require prior knowledge of the data which in practice implicitly or explicitly restrict them to a specific domain (as opposed to being domain agnostic)~\cite{CDP_survey_2017, FLUSS, CPD_human_motion_2016}. Another common limitation is that many CDPs algorithms are defined only for offline applications (i.e. they require having access to the full time series before performing the segmentation)~\cite{CDP_survey_2017, Offline_CDP_2020}. Further, for many real-world applications fast change-point detection are necessary to execute time-sensitive actions. Thus, algorithms that cannot handle online streaming data are ill-adapted for this type of reality, as they require a complete rerun every time a new data point is added. This problem is compounded by the fact that time series are often recorded over a long time and with a high sampling rate. Thus, algorithms with high computational complexity or poor scalability are also inappropriate for these types of real-world applications.
Therefore, developing an unsupervised semantic segmentation algorithm that is hyperparameter lite, domain agnostic, scalable and works on online streaming data is an auspicious target.

One state-of-the-art algorithm, which meets many of these requirements is: Fast Low-cost Unipotent Semantic Segmentation~\cite{FLUSS} (FLUSS). FLUSS is a domain agnostic, scalable CDP algorithm, that can work with both online and offline data. The main hypothesis behind the development of FLUSS is that subsequences (small snippets of the time series data extracted around each time step in the time series) in similar segments are more similar than subsequences occurring after a changepoint. FLUSS handles multidimensional data by implicitly calculating the likelihood of a change-point at each time step for each channel independently before taking the average likelihood over all the channels. However, some of the considered dimensions might not contain information that is helpful for the segmentation, or they might be heavily correlated. Thus, when taking the average over all the channels, these “not so useful” channels will dilute the information from the meaningful channels. The authors acknowledge this problem and recommend that when dealing with high dimensional data one should do a search or manually find the most useful subset of channels and remove the rest. 

As time series data is often sampled from multiple sources and sensors, especially when working with wearable devices, it would be helpful to have an algorithm that can automatically extract the most useful information from high-dimensional times series data during segmentation. Consequently, the main contribution of this work is the introduction of Latent Space Unsupervised Semantic Segmentation (LS-USS). LS-USS is an unsupervised semantic segmentation algorithm that is hyperparameter lite, domain agnostic and capable of working with both streamed online and offline multidimensional data. LS-USS is based on FLUSS, but instead of using the similarity between subsequences, it finds the similarity between lower-dimensional encodings of the multidimensional subsequences obtained via an autoencoder. The idea is that the dimensionality reduction learned by the autoencoder will reduce redundant and correlated information between channels, which in turn will improve time series segmentation.

This work is organized as follows; an overview of the related work is given in Section~\ref{sec:new_related_work}. Section~\ref{sec:preliminaries} introduces the notions necessary for the description of the proposed algorithm in Section~\ref{sec:method}. The experiments are then presented in Section~\ref{sec:experiments}. Finally, the results and their associated discussion are covered in Section~\ref{sec:results} and \ref{sec:discussion} respectively.

\section{Related Work}
\label{sec:new_related_work}

CPD algorithms can be divided into two categories depending on their reliance on labeled data: Supervised and Unsupervised CPD. Supervised CPD usually entails extracting subsequences over a sliding window where each subsequence is assigned a class or label. Most of the classifiers used in other machine learning tasks are also leveraged for supervised CPD such as naïve Bayes \cite{reddy_using_2010}, Support Vector Machines (SVMs) \cite{reddy_using_2010}, Gaussian Mixture Models (GMMs) \cite{reddy_using_2010}, Decision Trees\cite{reddy_using_2010}, Hidden Markov Models \cite{cleland_evaluation_2014}, and Neural Networks \cite{ronneberger_u-net_2015}. When labels are available, the supervised methods are often the preferred solution. Unfortunately, labeling time series data can be prohibitively expensive (in terms of time, cost and/or human labor) making it an impractical solution for a wide variety of domains. Another challenge with supervised CPD is that they generally require training samples from all possible states (classes) of the signals beforehand and thus cannot easily adapt to novel behavior in the time series. Therefore, unsupervised CPD remains of great interest in many practical applications. 

Most of the existing unsupervised CPD make statistical assumptions on the data (e.g. stationarity, independent and identically distributed)~\cite{kuncheva_change_2013, adams_bayesian_2007, malladi_online_2013, harchaoui_kernel_2008, rosenbaum_exact_nodate, friedman_multivariate_1979, itoh_change-point_2010}, and/or require extensive tuning of model parameters as they rely on predefined parametric models~\cite{kawahara_change-point_2007, itoh_change-point_2010, yamanishi_unifying_2002}. As such, the type of algorithms can be cumbersome to apply on new domains and less robust to changes in the data over time. In contrast, LS-USS was designed to work using only minimal assumptions on the data and without requiring domain knowledge.


Another common limitation of existing CPD algorithms is that they require being used on batched data and are thus ill-suited for real-time applications~\cite{CDP_survey_2017, rakthanmanon_time_nodate, kawahara_change-point_2007, itoh_change-point_2010} such as detecting change-points in a patient's vital signs~\cite{yang_adaptive_2006} or continuously monitoring the wear and tear of industrial robots~\cite{nentwich_combined_2021}.
In contrast, LS-USS can be efficiently updated every time a new data point is added to the time series allowing it to run in real-time.

A popular form of unsupervised CPD algorithms relies on clustering subsequences of the time series~\cite{yairi_fault_2001, fu_pattern_2001, li_malm_1998}. These methods cluster individual subsequences extracted from running a sliding window over the time series. The idea being that if two temporally close subsequences belong to different clusters, a change-point most likely exist between the two. However, Keogh et al.~\cite{keogh_clustering_nodate} have shown that using subsequence clustering essentially produces cluster centers which tend towards a sinusoidal signal with random phases that average to the mean of the time series for any dataset used. In other words, the change-points detected are essentially random. To address this issue, Keogh et al.~\cite{keogh_clustering_nodate} proposes to instead cluster the time series data points using \textit{motifs}. In the context of data mining, motifs can be seen as fingerprints for time series data, as all non-random time series data produced by the same process are bound to contain some reoccurring patterns. Thus, the difference between clustering motifs and subsequence clustering is that in the former case clusters are made up of short individual time series, while in the latter case the clusters are derived from subsequences extracted from a sliding window. In other words, clustering motifs means that the cluster-centers do not represents averages over all the data but averages over motifs (similar patterns in the data). Several methods have been proposed to extract motifs in time series data~\cite{mueen_exact_2009,rakthanmanon_time_nodate, bailey_streme_2021, matrix_profile_2016}. In particular, \textit{matrix profile}~\cite{matrix_profile_2016} is a recent approach to efficient motif discovery. As matrix profiles are a core aspect of LS-USS, they will be presented in more details in section~\ref{sec:matrix_profile}.

\begin{figure}[!htbp]
    \centering
    \includegraphics[width=\linewidth]{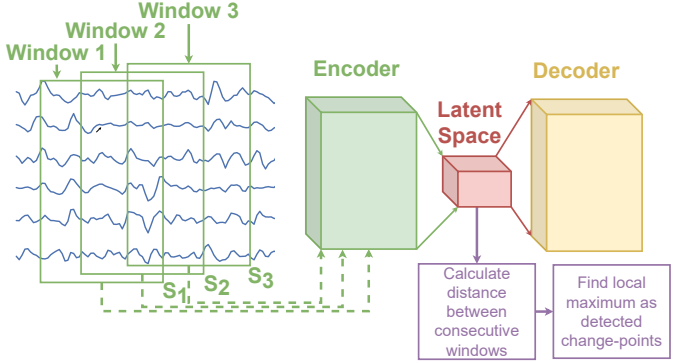}
    \caption{The diagram of the LFMD algorithm presented in~\cite{LFMD}. The distance of the latent features are calculated between each adjacent window. The distance's local maxima are selected as the change-points to be returned by the algorithm.5}
    \label{fig:LFMD_diagram}
\end{figure}

Many of the current states-of-the-art CPD algorithms \cite{alippi_change_nodate, gu_unsupervised_2010, kawahara_sequential_2012, kawahara_change-point_2007}, including FLUSS~\cite{FLUSS}, were primarily designed for one-dimensional data. Unfortunately, this also means that they often cannot efficiently be applied to multidimensional data~\cite{CDP_survey_2017} as they poorly (if at all) take into account that multidimensional time series often have varying levels of correlated information across dimensions which can increase the complexity of finding change-points accurately. To alleviate this issue, CPD algorithms have been specifically designed for multidimensional data~\cite{qahtan_pca-based_nodate, kim_representation_2019, LFMD, sakurada_anomaly_2014, zhou_anomaly_2017, zhang_deep_2019}. In~\cite{qahtan_pca-based_nodate} principal component analysis (PCA) is first used to obtain a one-dimensional signal (using the principal component) before performing the segmentation. Kim et al.~\cite{kim_representation_2019} aims to segment out driving patterns from sensors on driving vehicles by leveraging word2vec~\cite{mikolov_distributed_nodate} to make an encoded representation of time series data consisting of both categorical and numerical information in varying scales. Autoencoders~\cite{autoencoder_review_2018} have also been employed in the domain of anomaly detection with great success~\cite{sakurada_anomaly_2014, zhou_anomaly_2017, zhang_deep_2019}. The most common approach is to first train the autoencoder on time series data which does not contain any anomalies (or as few as possible). At inference time, the reconstruction error is then used as a measure of how likely the current data contain an anomaly. The idea being that if the autoencoder cannot reconstruct the current signal well enough, it is most likely because it differs from what was seen during training. While this approach successfully finds change-points that are anomalies, it is not straightforward to adapt it for segmentation as the different segments are a natural part of the data. In other words, an autoencoder trained on specific segments will also have a low reconstruction error whenever the data is part of one of the trained segment. Nevertheless, as shown in~\cite{LFMD}, unsupervised segmentation can still be performed using an autoencoder while achieving state-of-the-art results by calculating a distance between consecutive window in the latent space. This method referred to as Latent Feature Maximal Distance (LFMD), is illustrated in Figure~\ref{fig:LFMD_diagram}. The central idea behind LFMD, using a learned latent space of an autoencoder as a way to efficiently characterize multidimensional data for CDP is also a core concept in LS-USS. As such, LFMD will be used in this work to better contextualize the performance of the proposed algorithm. 

\section{Preliminaries}
\label{sec:preliminaries}
The following section presents an overview of the fundamental building blocks used by LS-USS.

\subsection{Matrix Profiles}
\label{sec:matrix_profile}

\begin{figure}[!htbp]
    \centering
    \includegraphics{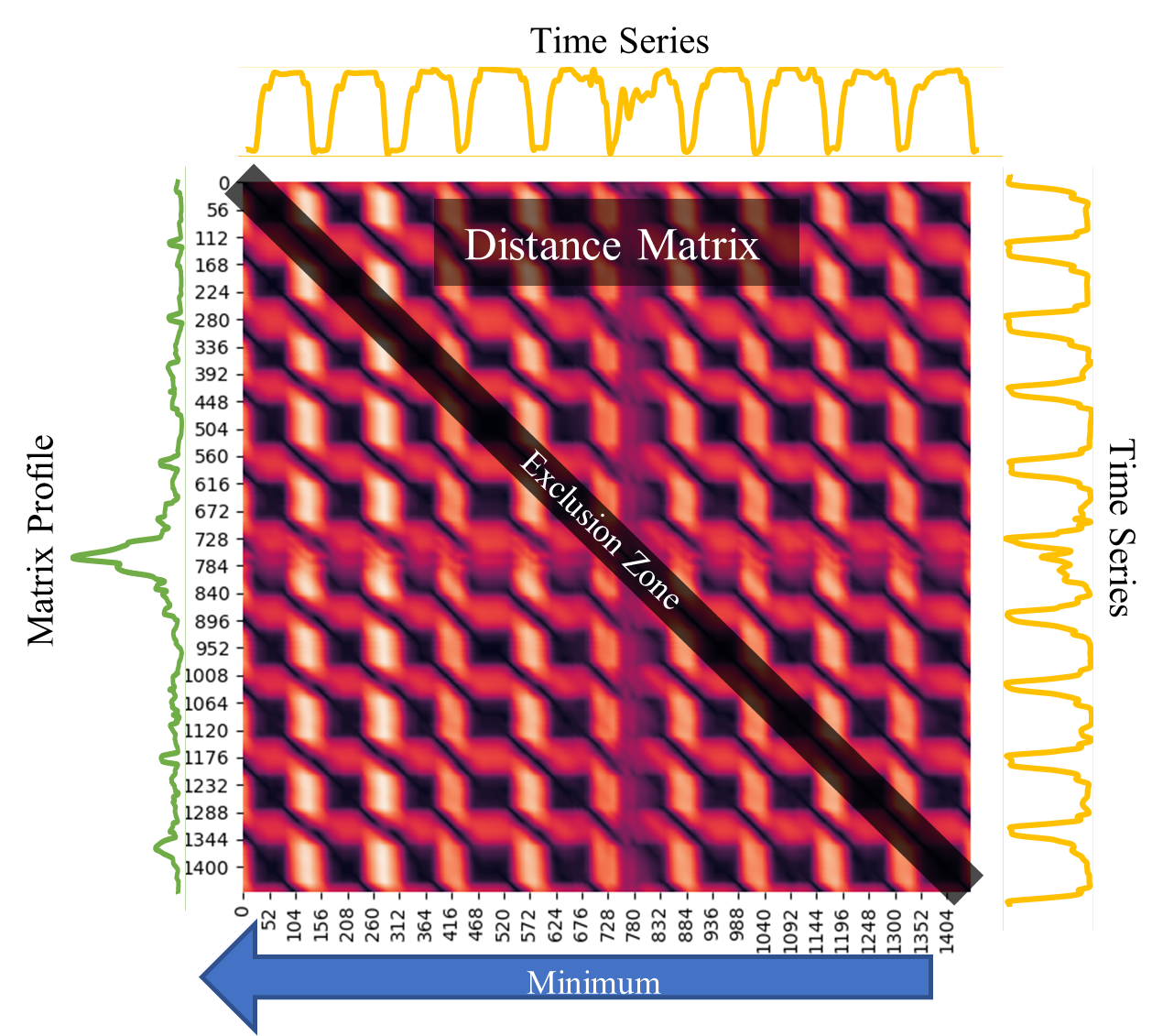}
    \caption{Plot depicting a time series and its corresponding distance matrix and matrix profile. The time series presented is an excerpt from the dataset NoGunGun from the UCR time series archive.}
    \label{fig:matrix_profile}
\end{figure}

One of the core concepts behind LS-USS is the matrix profile~\cite{matrix_profile_2016}, a data structure for time series that facilitates change-point detection and motif discovery. The matrix profile represents the distance between each motif and their most similarly associated motif (excluding themselves). To calculate the matrix profile, one first has to find the set of all subsequence \textbf{A} from the time series \textbf{T} by utilizing a sliding window of length $m\in\mathbb{N}$ using a step size of 1 to extract all the possible subsequences of \textbf{T}. After this a distance matrix is calculated by computing the z-normalized Euclidian distance between every subsequence in \textbf{A} with every other subsequence in \textbf{A}. Figure~\ref{fig:matrix_profile} shows what the resulting distance matrix looks like for the time series GunPoint from the UCR time series archive~\cite{UCRArchive2019}. Each row in the distance matrix, referred to as the distance profile \textbf{D} depicts the distance from the subsequence at the current row index to every other subsequence in the time series. The distance profile \textbf{D} is calculated efficiently in $O(nlog(n)$, using a technique referred to as Mueen’s algorithm for similarity search (MASS)~\cite{rakthanmanon_searching_2012}.

The matrix profile is then obtained by taking the Euclidian distances between all the subsequences in \textbf{A} with the nearest non-trivial neighbor in \textbf{A} itself. In other words, the matrix profile is created by extracting the smallest value from each row in the distance matrix, as this will be the distance to the nearest neighboring subsequence. Note, however, that the most similar subsequence to a given subsequence will always be the subsequence itself (as they are identical). Similarly, subsequences that are extracted close in time will also be nearly identical. To avoid these trivial matches, an exclusion zone around each index is created by setting the values in these areas of \textbf{A} to infinity. As seen in Figure~\ref{fig:matrix_profile}, this exclusion zone will be along the diagonal of the distance matrix. 

\begin{figure}[!htbp]
    \centering
    \includegraphics[width=\linewidth]{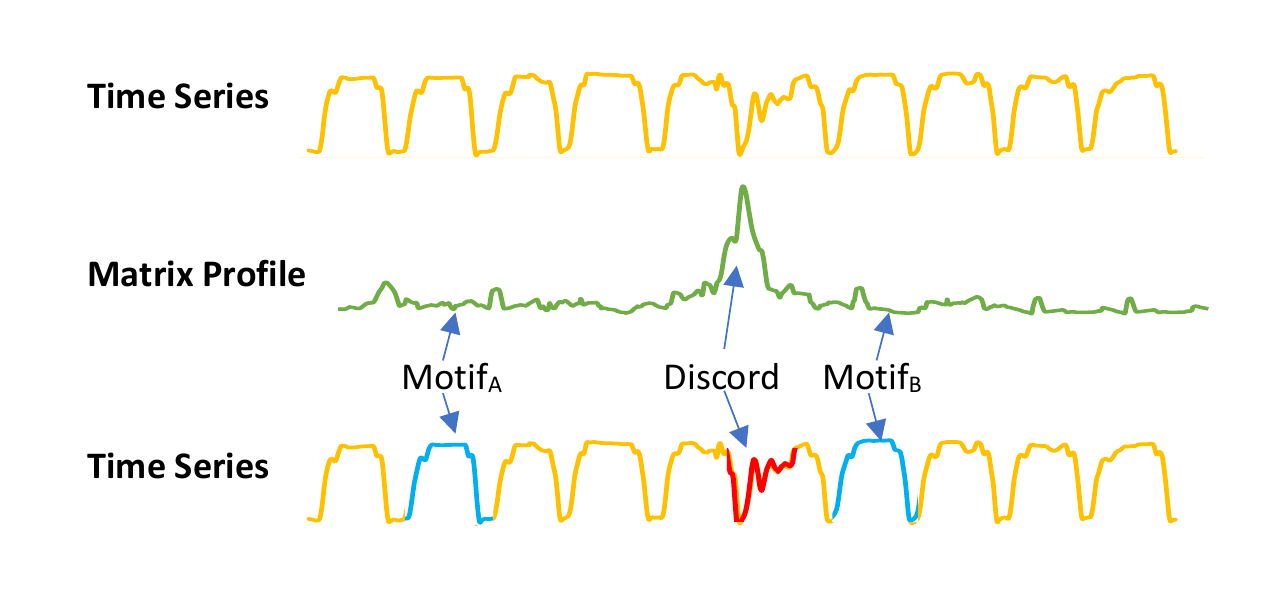}
    \caption{The distance matrix is a matrix constructed from all the distance profiles, which shows how similar each subsequence in the times series is to every other subsequence in the same time series. The matrix profile is derived by taking the minimum distance of each row. The original time series is shown in yellow and the matrix profile in green. }
    \label{fig:motif_vs_discord}
\end{figure}

The matrix profile can be used to facilitate the discovery of motifs and discord in the data. In the areas with relatively low values, the subsequences in the original time series must have (at least one) relatively similar subsequence elsewhere in the data. These regions are reoccurring patterns which are classified as motifs and always come in pairs. In contrast, for areas with relatively high values, the subsequence in the original time series must be a unique shape since it does not match any other subsequence. These areas of discords can be considered anomalies in the data. Figure~\ref{fig:motif_vs_discord} shows an example of how motifs and discords can easily be identified using a matrix profile.

When calculating the matrix profile $\vec{p}_A$, one can also extract the index of the nearest neighbor for each row in the distance matrix to make the matrix profile index $\vec{i}_A$, which is defined as the vector containing the indices of the nearest non-trivial neighbor in \textbf{A} for every subsequence in \textbf{A}. The matrix profile index can thus be seen as a time series containing pointers to the nearest neighbors for each subsequence and will serve as the basis for the segmentation algorithm in LS-USS. The matrix profile can be efficiently calculated in $O(n^2 log(n))$ using the Scalable time series Anytime Matrix Profile (STAMP)~\cite{matrix_profile_2016} outlined in Algorithm~\ref{alg:STAMP}.


\begin{algorithm}[!htbp]
\caption{Scalable time series Anytime Matrix Profile (STAMP)}
\label{alg:STAMP}
\textbf{Input}: \\
\textit{$\vec{t}$ – time series} \\
\textit{m - subsequence length} \\
\textbf{Output}: \\
\textit{$\vec{p}_{A}$,$\vec{i}_{A}$ – The incrementally updated matrix profile and the associated matrix profile index}
\begin{algorithmic}[1]
\Procedure{STAMP}{\textit{$\vec{t}$,m}}
\State \textit{$\vec{p}_{A}$ = inf's, $\vec{i}_{A}$ = zeros}
\State \textbf{A} $\gets $\textit{the all-subsequence set from $\vec{t}$}
\For{\textit{idx=0 : length($\vec{t}$)-m}}
    \State \textit{D[idx] = MASS(\textbf{A}[idx],TA) \Comment{MASS calculates the distance profile for the current subsequence}}
    \State \textit{$\vec{p}_{A}$[idx],$\vec{i}_{A}$[idx] = ElementWiseMin(\textbf{D}[idx]) Save the minimum values in the distance profile and the corresponding index}
\EndFor
\Return $\vec{p}_{A}$,$\vec{i}_{A}$
\EndProcedure
\end{algorithmic}
\end{algorithm}

\subsection{Fast Low-cost Unipotent Semantic Segmentation}
\label{sec:fluss}
Fast Low-cost Unipotent Semantic Segmentation (FLUSS)~\cite{FLUSS} is a segmentation algorithm that builds upon the matrix profile. This algorithm utilizes the matrix profile index ($\vec{i}_{A}$), to segment time series data. 
The intuition behind FLUSS can be illustrated through this simple example: 
In time series, data gathered from a person wearing an activity sensor and performing the activities walking and running, one would expect that most of the walking subsequences would point to other walking subsequences, and most running subsequences would point to other running sequences. In other words, for a given index in the time series $\vec{t}$, the number of arcs crossing "over" that index would be small if it is in an area where the time series is changing, whereas a high arc-count would be expected in areas with a clear homogeneous pattern. If these arc-crossings are counted for each index, the end result is the Arc Curve (AC).

The AC of a given time series $\vec{t}$ of length $n$ will itself be a time series of length $n$ containing only non-negative values. The value at the $i^{th}$ index in the AC specifies the number of arcs spatially crossing over location $i$ in the original time series $\vec{t}$~\cite{FLUSS}. 

\begin{figure}[!htbp]
    \centering
    \includegraphics[width=\linewidth]{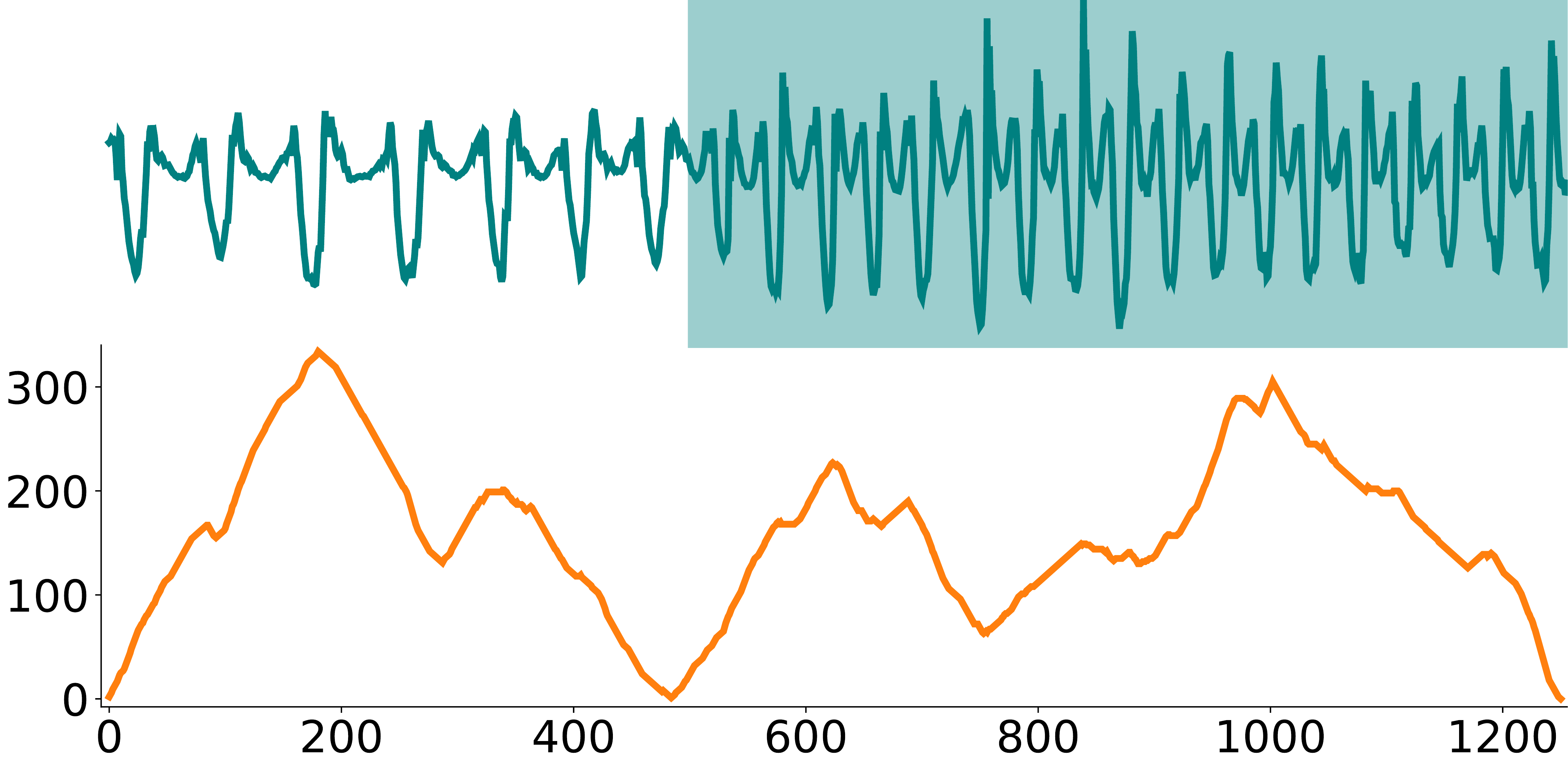}
    \caption{The top plots show a time series recorded from a gyroscope placed on a person's arm~\cite{walking_vs_jogging}. During the recording, the person is initially walking but then starts jogging at around timestep 500. The bottom plot shows the corresponding arc curves. Importantly, the arc counts are low around timestep 500, corresponding to the time series true change-point. However, the arc counts also decrease towards the beginning and end of the time series, which corresponds to a construction artifact that has to be corrected for.}
    \label{fig:CAC}
\end{figure}

\begin{figure}[!htbp]
    \centering
    \includegraphics[width=\linewidth]{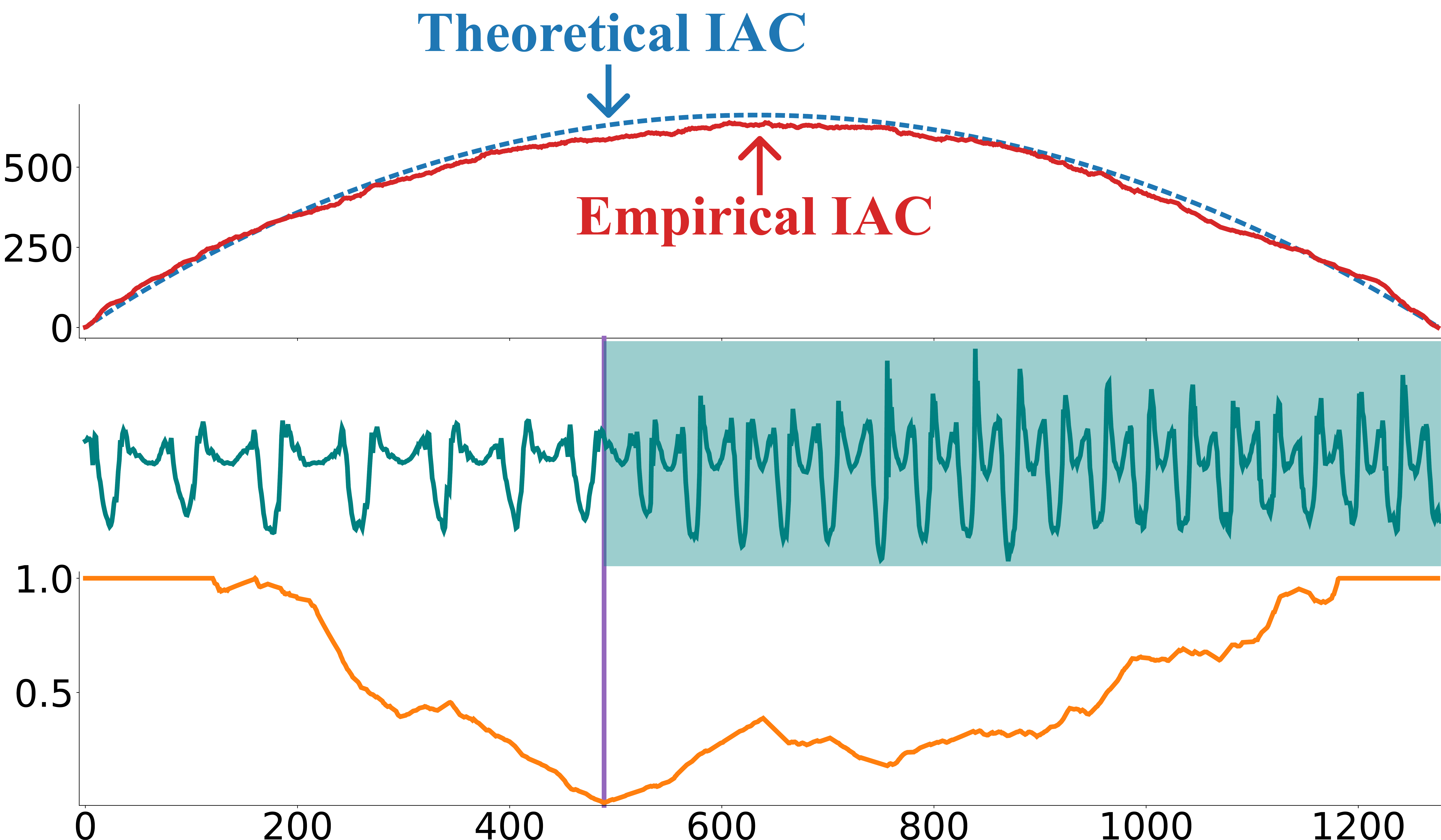}
    \caption{The top plot shows the empirical vs theoretical idealized arc curve (IAC). The bottom plot shows the corrected arc curve (CAC) computed on the walking-jogging time series from~\cite{walking_vs_jogging}. As shown by the purple vertical line, the minimum value on the CAC can be used to identify this time series' change-point.}
    \label{fig:CAC_IAC}
\end{figure}

Figure \ref{fig:CAC} shows that the AC is close to zero at the transition between the segments. However, by definition, the arc count will naturally be lower closer to the edges until it become zero as no arcs can cross the borders of the time series.
To compensate for this, the authors in~\cite{FLUSS} divide the AC with an inverted parabola with a height equal to half the length of the time series. This parabola is called the idealized arc curve (IAC) and is what the AC would look like for a time series with no structure, where all arc curves would just point to random locations. The empirical and theoretical IAC is depicted in the top plot in Figure \ref{fig:CAC_IAC}. Dividing the AC with the IAC will normalize the time series between 0 and 1 and solve the edge effects. The resulting vector is known as the Corrected Arc Curve (CAC). A min function is also applied to ensure that the CAC is between 0 and 1, even in the unlikely event that AC $>$ IAC.

\subsubsection{Regime Extraction Algorithm}
Low values in the CAC are used to identify potential change-points. In~\cite{FLUSS}, change-points are then selected using the Regime Extraction Algorithm (REA). REA search for the $k$ lowest “valley” points in the CAC. 
However, if the point at time $t$ is the lowest point on the CAC, the points at time $t-1$ and $t+1$ are likely the second and third-lowest point. To avoid all the change-points ending up in one “valley”, an exclusion zone is set around the already detected “valley” points, the width of which is an hyperparameter. Note that REA only works when the number of change-points is know beforehand. Unfortunately such a requirement can be hard/impossible to fulfill, in practice, for a wide variety of applications. The pseudo-code for REA is outlined in Algorithm~\ref{alg:REA}.


\begin{algorithm}[!htbp]
\caption{Region Extraction Algorithm (REA)}
\label{alg:REA}
\textbf{Input}: \\
\textit{CAC - a Corrected Arc Curve} \\
\textit{numRegimes - number of regime changes} \\
\textit{NW - Subsequence size} \\
\textbf{Output}: \\
\textit{locRegimes - the location of the change-points}
\begin{algorithmic}[1]
\Procedure{REA}{\textit{CAC,numRegimes, NW}}
\State \textit{locRegimes = empty array of length numRegimes}
\For{\textit{idx=0 : numRegimes}}
    \State \textit{locRegimes[i] = indexOf(min(CAC))}
    \State \textit{Set exclusion zone of 5 $\times$ NW around locRegimes[i] \Comment{To prevent matches to "self"}}
\EndFor
\Return \textit{locRegimes}
\EndProcedure
\end{algorithmic}
\end{algorithm}

\subsubsection{Fast Low-cost Online Semantic Segmentation}
\label{sec:FLOSS}

\begin{figure}[!htbp]
    \centering
    \includegraphics[width=\linewidth]{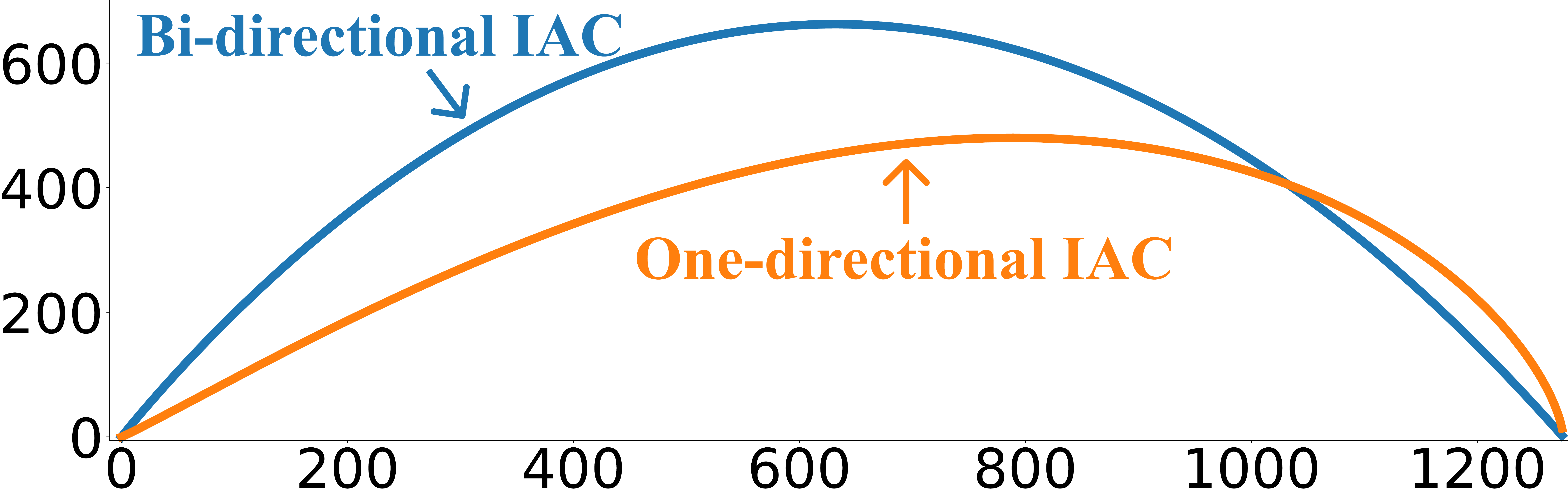}
    \caption{The bi-directional idealized arc curve (IAC) vs. the one-directional IAC.}
    \label{fig:1D_IAC}
\end{figure}

Adding a new point to the CAC takes only O($n log(n)$). However, removing the oldest point takes O($n^2$) as every subsequences could point to the ejected point, thus requiring the whole matrix profile to be updated. As such, while FLUSS can easily be run on offline datasets, it does not scale well in the context of streaming data. To solve this issue, the authors in~\cite{FLUSS} propose an online version of FLUSS, referred to as Fast Low-cost Online Semantic Segmentation (FLOSS). FLOSS addresses the online streaming issue by explicitly forcing each arc to only point towards a newer data point. Consequently, as no arc can point to an older point, ejecting the oldest point in the currently considered time series can now be done in O($1$). Thus, maintaining the one-directional CAC$_{1D}$ can be done in O($nlog(n)$). It should be noted however that in the case of the CAC$_{1D}$, the IAC will now be skewed to the right as the rightmost part of the CAC$_{1D}$ will have a higher chance of arc crossings (see Figure \ref{fig:1D_IAC}).

FLOSS in conjunction with the one-directional CAC thus enables online streaming using the matrix profile. However, it should be expected that FLOSS will perform worse than FLUSS due to the directional constraints imposed on the arcs.

\subsubsection{Locality - Temporal Constraint}
\label{sec:TemporalConstraint}
In some time series datasets, the same activities or events can arise multiple times in a disjointed manner (e.g. a time series representing the electrocardiogram of a person laying down and standing up multiple times in succession). In these cases, the CAC is not a good indicator of a regime change as the arcs for a given event-type (e.g. laying down) would have practically the same likelihood of pointing to subsequences in the segment they are occurring in than they would have in pointing to subsequences from other segments of the same event-type.
This challenge can be addressed by using a temporal constraint (TC) that limits how far in time an arc can point, thus forcing each index in the CAC to only consider a local area around itself. While this adds a hyperparameter (length of TC) to the algorithm, it also reduces the computational time of FLUSS. If considered, this hyperparameter is suggested in~\cite{FLUSS} to be set to circa maximum expected segment length, thus allowing to inject some domain knowledge to the algorithm if available. Furthermore, as pointed out in~\cite{FLUSS}, this hyperparameter is not very sensitive and thus only requires a rough idea of the order of the temporal scale at which the change-points occurs.



\subsection{Autoencoder}
\label{sec:cpd_autoencoder}
Autoencoders~\cite{autoencoder_review_2018} are a type of neural network which aim to learn a mapping from a high-dimensional observation to a lower-dimensional feature space (latent space) in an unsupervised manner. Importantly, this mapping is learned with the aim of being able to reconstruct as closely as possible the original observation from the lower-dimensional feature space.

In other words, define $\mathcal{X}$ as the input space and consider a probability measure $p$ on $\mathcal{X}$. The goal of an autoencoder is then to learn two functions $\psi$ and $\phi$, such that: 
\begin{align}
    \psi: \mathcal{X} \rightarrow \mathcal{Z}\\
    \phi: \mathcal{Z} \rightarrow \mathcal{X}\\
    \psi, \phi = \operatorname*{arg\,min}_{\psi, \phi} \mathds{E}_{x \sim p}\Big[\parallel x - (\phi \circ \psi)(x)\parallel\Big]
\end{align}

Where $\mathcal{Z}$ is the latent feature space that serves as an information bottleneck to the original feature space.

The input of the autoencoder is the time series partitioned into windows of length $w$. Note that these subsequences can be defined with an overlap. The detailed autoencoders' architectures employed in this work will be presented in Section~\ref{sec:method}.



\section{Method}
\label{sec:method}

LS-USS proposes to perform unsupervised segmentation by leveraging an autoencoder to learn a meaningful latent feature space from which the corrected arc curve can be computed from. The following details the proposed algorithms and contributions of this work. 

\subsection{Autoencoder Implementation}
For comparison's sake, the fully connected network's architecture used by LFMD~\cite{LFMD} is employed in this work. Further, a simple convolutional network is also considered to evaluate if modeling the input's temporal characteristics can help find a better latent representation than the fully connected version.
\subsubsection{Fully Connected Model}
Following the autoencoder implementation~\cite{LFMD}, the fully connected model reproduced uses two hidden layers for both the encoder and decoder, transposed weights for the decoder, and the sigmoid function as its activation function. For the decode encoder, each hidden layer is half the size of the previous layer and the opposite for the decoder layer (the initial size of which will depend on the input/subsequence size). Adam~\cite{kingma_adam_2017} is employed to optimize the network's weights using the mean square error as the loss function. Finally, the dimension of the latent space is defined so that the ratio between the feature representation and the input $\frac{dim(x)}{dim(z)}$ is 0.1. Where $x\in\mathcal{X}$ and $z\in\mathcal{Z}$. 

\subsubsection{Convolutional Model}
The architecture employed for the convolutional model is presented in Table~\ref{table:convolutional_network_architecture}. The model is trained in the same way as the fully connected model using mean square error as loss function and Adam as optimizer. The feature size, which is now defined as the ratio ($\frac{dim(z)}{NC\times NW}$), is set to $\frac{1}{6}$.

\begin{table*}[!htbp]
\begin{center}
\begin{tabular}{@{}cccccccccc@{}}
\toprule
Layer & Type & Input Length & Output Length & Input Width & Output Width & Kernel Size & Stride & Padding & Activation \\ \midrule
\begin{tabular}[c]{@{}c@{}}Hidden Layer 1\\ Encoder\end{tabular} & Convolution & NW & NW/2 & NC & 2xNC & 3 & 2 & 1 & ReLU \\
\begin{tabular}[c]{@{}c@{}}Hidden Layer 2\\ Encoder\end{tabular} & Convolution & NW/2 & NW/4 & 2xNC & 4xNC & 3 & 2 & 1 & ReLU \\
\begin{tabular}[c]{@{}c@{}}Reshape\\ Encoder\end{tabular} & Reshaping & NW/4 & NW & 4xNC & 1 & - & - & - & - \\
\begin{tabular}[c]{@{}c@{}}Hidden Layer 3\\ Encoder\end{tabular} & Fully Connected & NW & NW/2 & 1 & 1 & - & - & - & ReLU \\
\begin{tabular}[c]{@{}c@{}}Hidden Layer 4\\ Encoder\end{tabular} & Fully Connected & NW/2 & NW/4 & 1 & 1 & - & - & - & ReLU \\
\begin{tabular}[c]{@{}c@{}}Hidden Layer 5\\ Encoder\end{tabular} & Fully Connected & NW/4 & NW/6 & 1 & 1 & - & - & - & ReLU \\
\begin{tabular}[c]{@{}c@{}}Hidden Layer 1\\ Decoder\end{tabular} & \begin{tabular}[c]{@{}c@{}}Transposed Hidden\\ Layer 5 Encoder\end{tabular} & NW/6 & NW/4 & 1 & 1 & - & - & - & ReLU \\
\begin{tabular}[c]{@{}c@{}}Hidden Layer 1\\ Decoder\end{tabular} & \begin{tabular}[c]{@{}c@{}}Transposed Hidden\\ Layer 4 Encoder\end{tabular} & NW/4 & NW/2 & 1 & 1 & - & - & - & ReLU \\
\begin{tabular}[c]{@{}c@{}}Hidden Layer 2\\ Decoder\end{tabular} & \begin{tabular}[c]{@{}c@{}}Transposed Hidden\\ Layer 3 Encoder\end{tabular} & NW/2 & NW & 1 & 1 & - & - & - & ReLU \\
\begin{tabular}[c]{@{}c@{}}Reshape\\ Decoder\end{tabular} & Reshaping & NW & NW/4 & 1 & 4xNC & - & - & - & - \\ 
\begin{tabular}[c]{@{}c@{}}Hidden Layer 1\\ Encoder\end{tabular} &  Transposed Convolution & NW/4 & NW/2 & 4xNC & 2xNC & 3 & 1 & 1 & ReLU \\
\begin{tabular}[c]{@{}c@{}}Hidden Layer 1\\ Encoder\end{tabular} & Transposed Convolution & NW/2 & NW & 2xNC & NC & 3 & 2 & 1 & Linear \\ \bottomrule
\end{tabular}
\end{center}
\caption{Overview of the architecture use for the convolutional network version of the autoencoder. NC is the original number of channels from the input data, while NW represents the sub-sequence length. 
}
\label{table:convolutional_network_architecture}
\end{table*}

\subsection{Latent Space Matrix Profile}

As previously stated, LS-USS leverages an autoencoder as a way to learn a meaningful representation of a multidimensional time series. Because the time series is first segmented into windows with each window going through the autoencoder, the resulting representation of the time series in the latent space is not a continuous time series. Unfortunately, this also means that the matrix profile cannot be computed from this latent space using the STAMP algorithm as presented in Section~\ref{sec:matrix_profile} as doing so is only possible if a sliding window can be applied to the time series itself. Thus, this work introduces the Latent Space Matrix Profile (LSMP) which, as the name implies, is the matrix profile computed directly on the latent representation. The LSMP is defined by the same logic used for calculating the regular matrix profile (see Section~\ref{sec:matrix_profile}). The difference being the use of latent representation instead of subsequences. By exploiting the fact that each distance calculation can be done independently, it is possible to implement the computation of the Euclidean distance between pairs of vectors to run in parallel (e.g. GPU). The pseudo-code to compute the lsmp is presented in Algorithm~\ref{alg:LSMP}.



\begin{algorithm}[!htbp]
\caption{Latent Space Matrix Profile (LSMP)}
\label{alg:LSMP}
\textbf{Input}: \\
\textit{$T_{A}$ – time series} \\
\textit{m - subsequence length} \\
\textbf{Output}: \\
\textit{$P_{F}$,$I_{F}$ – The incrementally updated latent space matrix profile and the associated latent space matrix profile index}
\begin{algorithmic}[1]
\Procedure{LSMP}{\textit{$T_{A}$,m}}
\State \textit{$P_{A}$ = inf's, $I_{A}$ = zeros}
\State $F$ $\gets $\textit{the latent all-subsequence set from $T_{A}$}
\For{\textit{idx=0 : length($T_{A}$)-m}}
    \State \textit{D[idx] = CDIST(A[idx],TA) \Comment{CDIST calculates the distance profile for the current subsequence.}}
    \State \textit{$P_{F}$[idx],$I_{F}$[idx] = ElementWiseMin(D[idx]) Save the minimum values in the distance profile and the corresponding index}
\EndFor
\Return $P_{F}$,$I_{F}$
\EndProcedure
\end{algorithmic}
\end{algorithm}

\begin{figure}[!htbp]
    \centering
    \includegraphics[width=\linewidth]{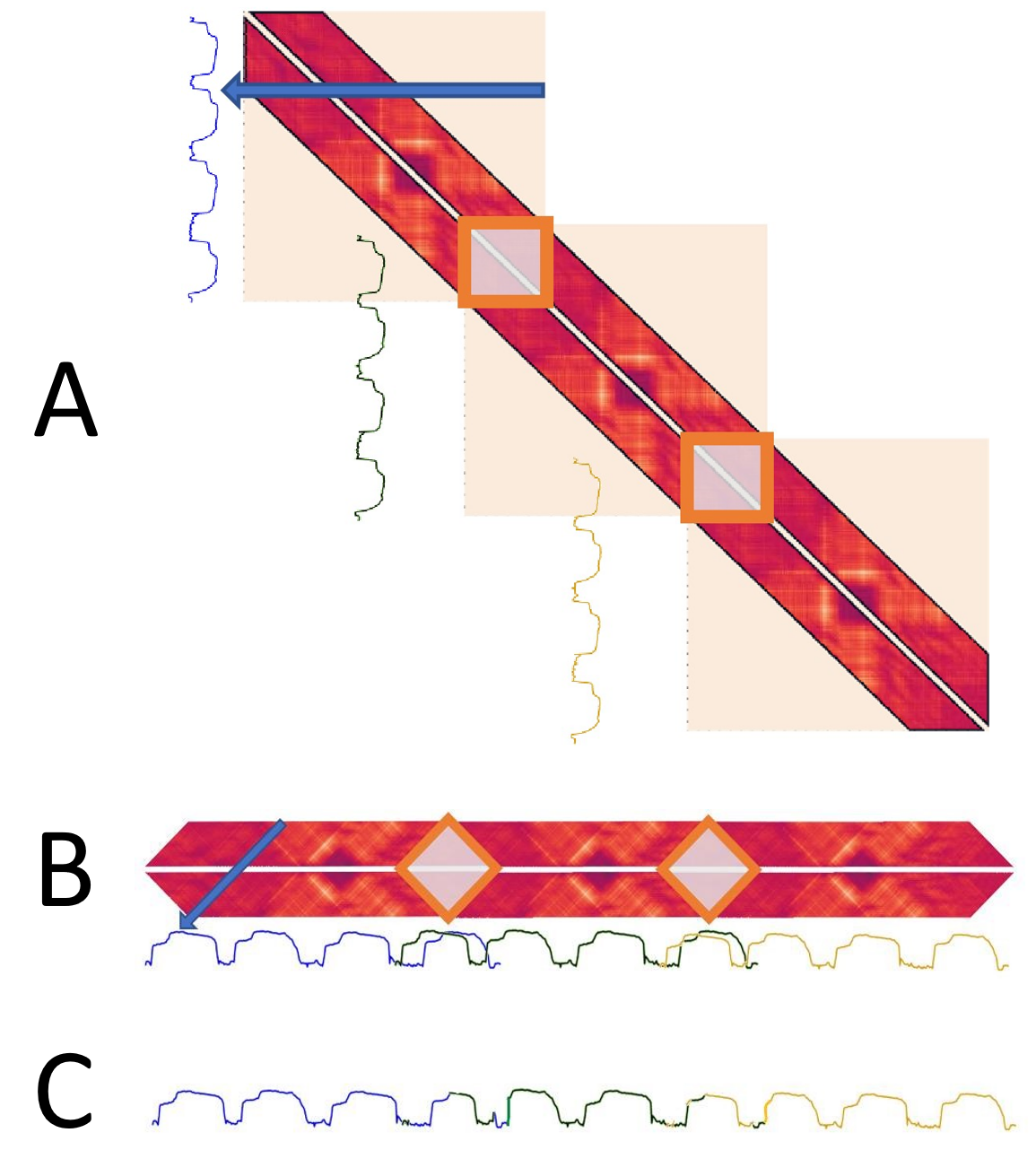}
    \caption{The top part of figure A (blue arrow) depicts the batched collapse of the temporally constrained latent distance matrix into the LSMP. Figure A as a whole shows the temporarily constrained matrix profile of a time series of length 10400. Figure B shows the three batched matrix profiles calculated where they overlap with each other. Figure C shows the full matrix profile after taking the minimum value in the overlapping area.}
    \label{fig:LSMP_illustration}
\end{figure}

Similarly to FLUSS, performing time series segmentation based on the LSMP and CAC is only meaningful when the same segment-type is not repeating over the time series (see Section~\ref{sec:TemporalConstraint}). 
Therefore, a TC is also applied when computing the CAC from the LSMP. 
Adding this TC also comes with the added benefit that only the distances between the latent features located inside the temporal constraint need to be calculated. The top of figure~\ref{fig:LSMP_illustration}-A depicts the collapse of the temporally constrained latent distance matrix into the LSMP.


A memory problem arise for long time series as it is necessary to save 2*TC data points per time step to make the temporally constrained distance profile, which can rapidly become intractable. 
To address this memory issue, instead of considering the full distance matrix (constrained by TC), this work instead propose to use an overlapping window on the time series where a distance matrix will be computed for each window. A matrix profile for each distance matrix is then computed. Finally, the different matrix profiles are merged together by taking the minimum value (and the corresponding index) over the overlapping area of each matrix profile. This operation is referred to as the batched collapse of the matrix profile, a depiction of which is shown in Figure~\ref{fig:LSMP_illustration} (the pseudo-code is also provided in Algorithm~\ref{alg:batched_collapse_algorithm}).





\begin{algorithm}[!htbp]
\caption{Batched Collapse algorithm}
\label{alg:batched_collapse_algorithm}
\begin{algorithmic}[1]
\Procedure{BatchedCollapse}{\textit{$t_{lim}$}}
\State $D$ $\gets$ \textit{The distance matrix, calculated from timestamp 0 to $t_{lim}$}
\State $P$ $\gets $\textit{The matrix profile obtained by collapsing $D$}
\State $t_{lim\_curr} \gets t_{lim}$
\While{$t_{lim} < len(T)$}
    \State $t_{lim\_prev} \gets t_{lim\_curr}$
    \State $t_{lim\_curr} \gets t_{lim\_curr} + t_{lim}$
    \State $D_{new}$ $\gets$ \textit{Calculate distance matrix from $t_{lim\_prev}$-$TC\times2-1$ to $t_{lim\_curr}$}
    \State Collapse $D_{new}$ to get the matrix profile $P_{new}$
    \State $P_{merged}$ $\gets$ \textit{Merge $P$ with $P_{new}$ \Comment{keep the minimum value between $P$ and $P_{new}$ where the two vectors overlap in time}}
    \State $P$ $\gets$ $P_{merged}$
\EndWhile
\Return $P$
\EndProcedure
\end{algorithmic}
\end{algorithm}

As seen in the Batched collapse algorithm, collapsing the matrix profile before processing the entire time series comes at the cost of recalculating the last (TC*2-1) time steps of the matrix profile. This recalculation is necessary as the first time steps in the new matrix profile can point back to the old ones (see Figure~\ref{fig:LSMP_illustration}).

The “batch collapse” algorithm can also be used to make LS-USS $\epsilon$ real-time (i.e. an algorithm that requires at least $\epsilon$ data points to detect a change-point. A completely online algorithm would then be 1 real-time) by accumulating a batch of data before adding it to the end of the matrix profile. Doing this entails a trade-off between batch size and calculation time as every time a batch is added, TC*2-1 time step must be recalculated.

\begin{figure}[!htbp]
    \centering
    \includegraphics[width=\linewidth]{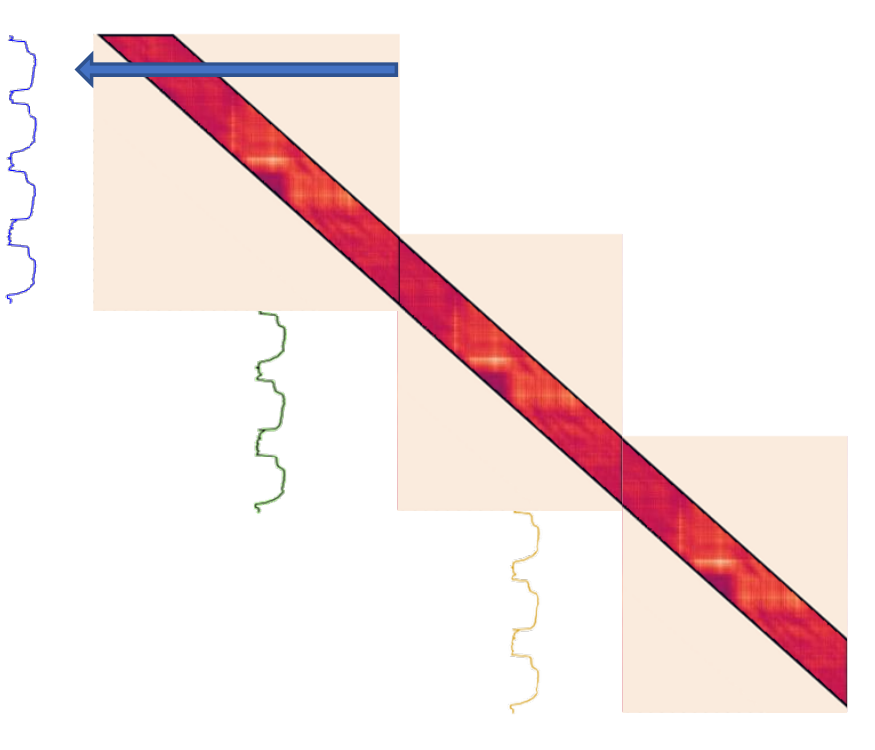}
    \caption{Online version of LSMP on a time series of length 10400. No recalculation is needed when updating the distance matrix}
    \label{fig:LSMP_online_illustration}
\end{figure}

In section~\ref{sec:FLOSS}, FLOSS was made to work online by forcing arcs to only be able to point forwards in time. This trick can also be borrowed to update the LSMP in real-time. If one only looks for the closest distance forward in time, no subsequence can have a nearest-neighbor in the previous LSMP. As such, the requirement of recalculating TC*2-1 time steps at each new batch also disappears (see Figure~\ref{fig:LSMP_online_illustration}). As with FLOSS, it should be expected that using only forward-pointing arcs will have a negative impact on the algorithm's performance.

\subsection{Latent Space Unsupervised Semantic Segmentation (LS-USS)}

LS-USS uses an autoencoder to encode the multidimensional all subsequences set \textbf{A} into the one-dimensional latent all subsequence set \textbf{F}. \textbf{F} is then used to compute the LSMP $\vec{P}_F$ and the corresponding LSMP index vector $\vec{I}_F$ as described in the previous section. The indices $\vec{I}_F$ are then used to make the CAC, which is the graph that contains the number of arc crossings at each time step. Like in FLUSS, when few arcs are crossings over a particular data point, this indicates a high likelihood of a change-point at that time step, and a high number of arc crossings indicates a low likelihood for a change-point.

\begin{figure}[!htbp]
    \centering
    \includegraphics[width=\linewidth]{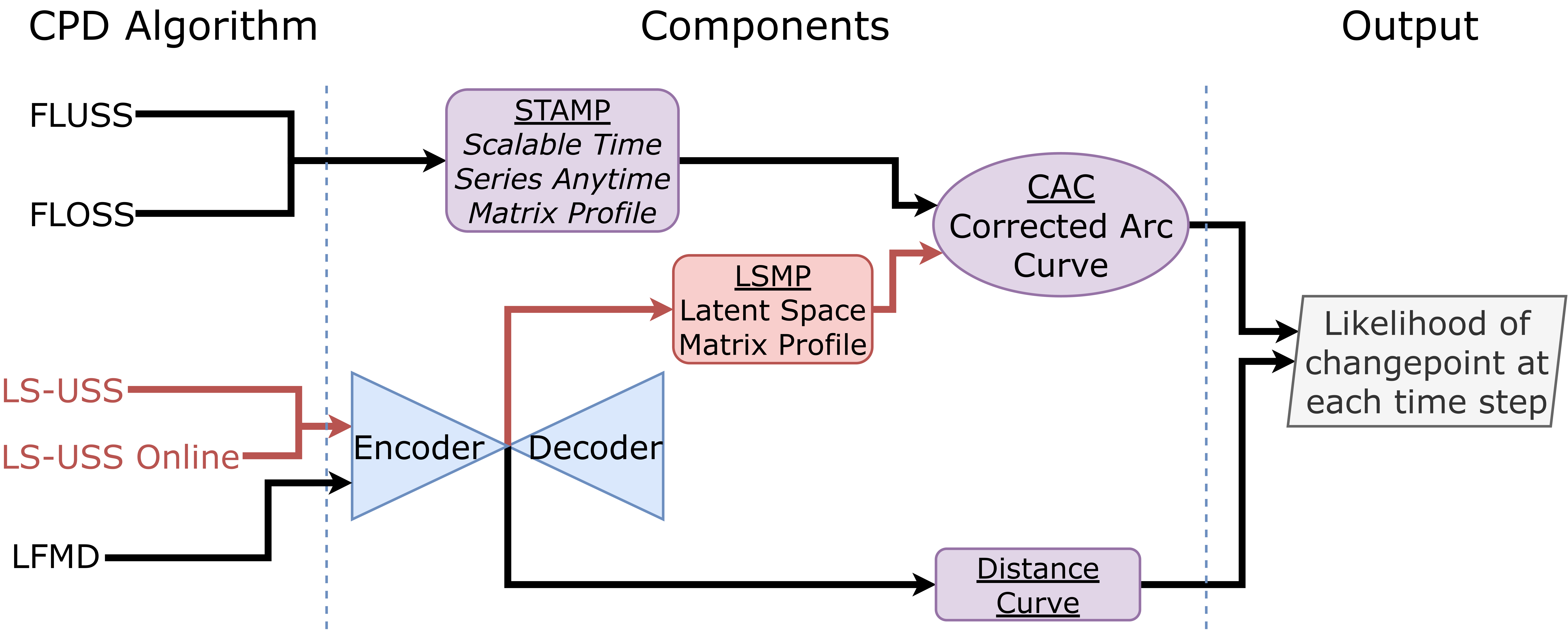}
    \caption{Overview of the LS-USS and LS-USS online algorithms compared to the other considered algorithms.}
    \label{fig:flowchart_algorithms}
\end{figure}

LS-USS online is similar to LS-USS, except that it uses the version of the LSMP that works online by only considering right-pointing arcs. Doing this makes it possible to update $\vec{I}_F$ without recomputing the distance matrix for the last ($TC\times 2-1$) time steps. As mentioned, the regular LSMP can also be $\epsilon$ real-time by accumulating a batch of data before adding it to the end of the $\vec{P}_F$, making the regular LS-USS $\epsilon$ real-time. An overview of the relation between LS-USS, LS-USS online, FLUSS, FLOSS and LFMD is shown in Figure~\ref{fig:flowchart_algorithms}. 

\subsection{Local Regime Extraction Algorithm}
REA was presented as an algorithm for extracting segments based on the k-lowest "valley" points of the CAC. For most long time series data the change-points are distributed relatively evenly in time. In these cases, when extracting change-points, local minimums are often more meaningful than the global minimums. The REA algorithm will always pick the global k-lowest "valley" points on the CAC instead of locating where the CAC is at its lowest compared to the local area around it. To address this issue, this work introduces the Local Regime Extraction Algorithm (LREA). The method scales each point in the CAC to zero mean and unit variance based on the local mean and standard deviation calculated over a rolling window (as shown in Equation~\ref{equation_lrea}). 
After the scaling, the same procedure used in the REA algorithm extracts the lowest k “valley points” from the scaled CAC.

\begin{equation}
\label{equation_lrea}
C A C_{\text {scaled }}=\frac{C A C-\mu_{\text {rolling }}}{\sigma_{\text {rolling }}}
\end{equation}

Note that by increasing the window size, the extraction of threshold will be increasingly global and thus tend towards REA. 

\subsection{Local Threshold Extraction Algorithm}

For comparisons between the offline CPD algorithms in this work, the algorithms REA and LREA are used. However, these algorithms require information about the number of segments, which is often unavailable in practice. To address this, this work introduces the Local Threshold Extraction Algorithm (LTEA). As the name suggests, this algorithm works by scaling the CAC before doing threshold-based change-point extraction.

The same CAC scaling used in the LREA algorithm is first applied to LTEA. Note that in the case where the algorithm is applied on an online (streaming) time series, the rolling statistics can naturally be computed only on prior data (which is the main difference with offline dataset). 

In LTEA, the scaled CAC-values over a given threshold are disregarded by being set to a value of one. When the scaled CAC is standardized to zero mean and unit variance, a good threshold value was found to be around minus one (one standard deviation).

\begin{figure}[!htbp]
    \centering
    \includegraphics[width=\linewidth]{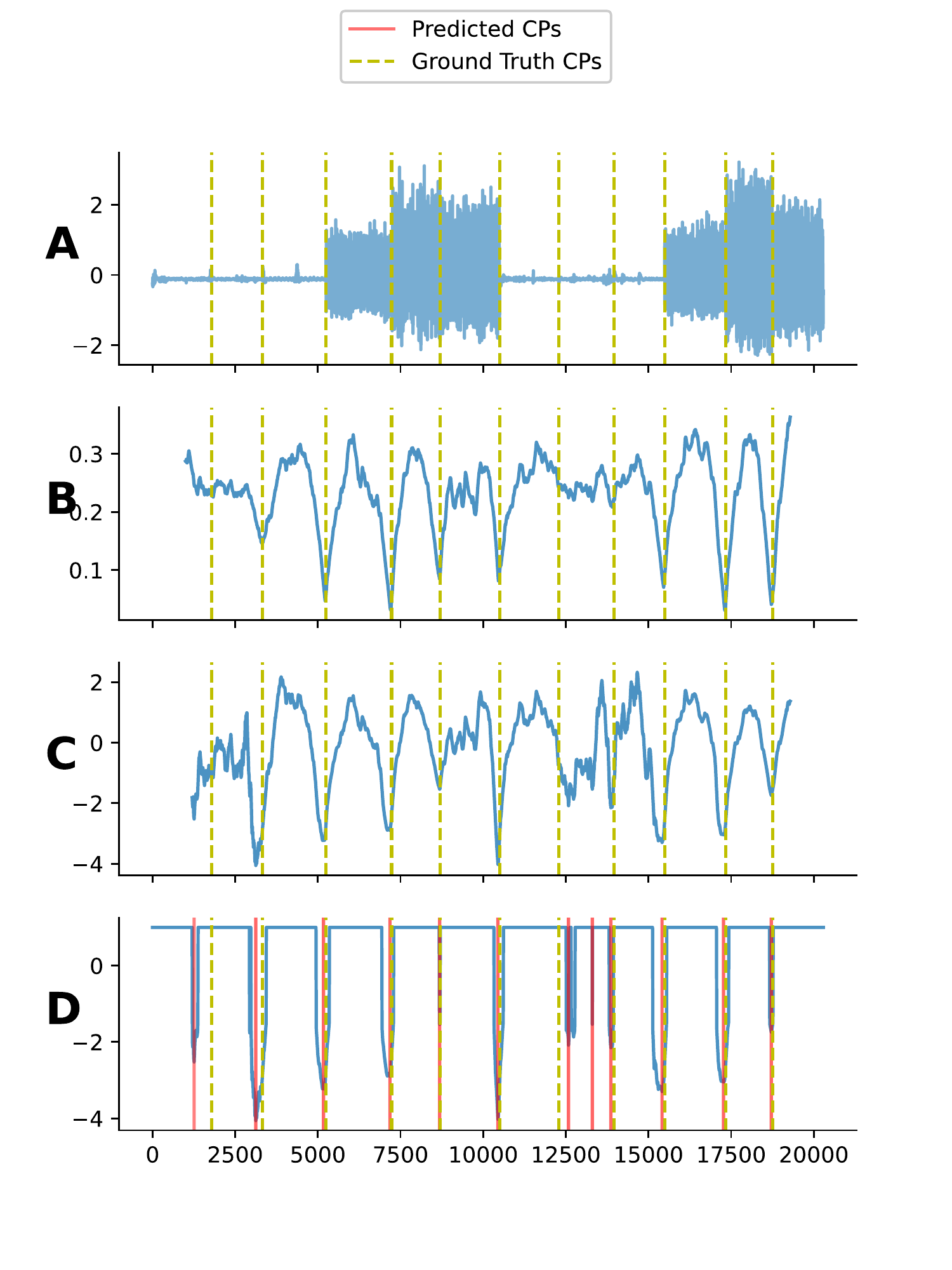}
    \caption{Plot A shows one channel from subject 4 in the UCI dataset. Plot B shows the regular CAC for subject 4. Plot C shows the scaled CAC. Plot D shows the thresholded CAC and the predicted change-points when using LTEA on this CAC. Note that as LTEA is not provided with the number of change-points in a given time series, it can in reality find more or less change-points than are actually present. In this case, an additional change-point compared to the ground truth was detected at index ca. 13500.}
    \label{fig:ltea}
\end{figure}

After thresholding the CAC, the local minimum obtained are identified as the change-point location. An exclusion zone like the one used in LREA and REA is also applied to avoid trivial matches caused by valleys close in time. The pseudo-code to compute LTEA is presented in Algorithm~\ref{alg:LTEA} and and example of the application of LTEA is shown in Figure~\ref{fig:ltea}.


\begin{algorithm}[!htbp]
\caption{Local Threshold Extracting Algorithm (LTEA)}
\label{alg:LTEA}
\textbf{Input}: \\
\textit{CAC – a Corrected Arc Curve or Distance Curve} \\
\textit{localWindowSize – The size of the window used to normalize the CAC} \\
\textit{threshold – Values of the $CAC_{scaled}$ above this threshold will be set to 1. A good default value is to set this to -1.} \\
\textbf{Output}: \\
\textit{locRegimes – the location of the change-points}
\begin{algorithmic}[1]
\Procedure{LTEA}{\textit{CAC, localWindowSize, threshold=-1.}}
\State \textit{$CAC_{scaled}$  $\gets $ empty array with same length as the CAC}
\For{i=0:\textit{length($CAC_{scaled}$)}}
    \State \textit{localWindow  $\gets $ CAC[idx-(localWindowSize) : idx+(localWindowSize)]}
    \State \textit{$\mu_{rolling}$   $\gets $ mean(localWidow) 
    }
    \State \textit{$\sigma_{rolling}$  $\gets $ std(localWindow) 
    }
    \State \textit{$CAC_{scaled}$[i]  $\gets $ (CAC[i] - $\mu_{rolling}$) / $\sigma_{rolling}$  
    }
    \If{\textit{$CAC_{scaled}$[i] $>$ threshold}}
        \State \textit{$CAC_{scaled}$[i]  $\gets $ 1}
    \EndIf
\EndFor
valleys $\gets $ List of sequences in $CAC_{scaled}$, where a sequence correspond to all the consecutive points with a value different than 1 in the $CAC_{scaled}$. 
\For{\textit{for valley in valleys}}
    \State \textit{locRegimes[i]  $\gets $ indexOf(min(valley))}
\EndFor
\Return locRegimes
\EndProcedure
\end{algorithmic}
\end{algorithm}

\section{Experiments}
\label{sec:experiments}
\subsection{Datasets}
\label{sec:dataset_section}
To benchmark the performance of LS-USS (and LS-USS Online) against FLUSS, FLOSS and LFMD, two biosignals-based dataset are considered. 
\subsubsection{UCI Human Activity Recognition}
The UCI Human Activity Recognition Using Smartphones Dataset~\cite{UCI_dataset} contains 30 volunteers between the ages of 19 and 48 who perform six activities wearing a smartphone on the waist. The activities are walking, walking up stairs, walking down stairs, sitting, standing, and laying down. The collected data came from the 3-axial accelerometer and 3-axial gyroscope located in a Samsung Galaxy S2 and were sampled at 50Hz. The raw acceleration data signals have three main components: body movement, gravity, and noise. The sensor data was filtered using a median filter and a 3rd order low pass Butterworth filter with a corner frequency of 20 Hz for noise removal. A low pass Butterworth filter, with a corner frequency of 0.3 Hz, is used to separate the body movement and gravity signals. The reason for choosing to use a 0.3 Hz cut-off frequency is that gravitational forces are assumed only to have low-frequency components. This work will use the data from the gyroscope, accelerometer, and the filtered body movement from the accelerometer. As all these signals are sampled in all three spatial dimensions, the dataset contains nine channels in total. Nine subjects constitute the training set, five subjects are used as a validation set, while the remaining 16 subjects are picked for the test set.
The labeled change-point detection is available in this dataset with the goal of detecting when the participant starts a new activity. This dataset will be referred to as the UCI dataset. 

\subsubsection{EMG-Based Long-Term 3DC Dataset}
The EMG-Based Long-Term 3DC Dataset~\cite{EMG_dataset} contains data from 20 able-bodied participants performing eleven hand gestures while recording their forearm's muscle activity over a period of 21 days (the recording sessions took place every $\sim$7 days). The goal for the unsupervised segmentation is to detect when a participant transitions towards a new gesture. This is of particular interest for myoelectric-based control as being able to detect such transitions accurately in real-time would improve the performance of previously proposed self-learning classifiers~\cite{emg_self_supervision}. 
The armband used when recording this dataset is the 3DC Armband~\cite{3dc_armband}. The 3DC is a ten-channel, dry electrode 3D printed EMG band with a sampling rate of 1000~Hz per channel. While the armband also features a 9-axis Magnetic, Angular Rate, and Gravity (MARG) sensor, only the 10 EMG channels are considered in this work. The dataset is divided into training sessions and evaluation sessions. In the training session, the participants were asked to hold each of the 11 gestures for 5 seconds. The transitions between gesture were not recorded, yielding a discontinuous time series. This procedure was repeated four times by each participant on every recording day. For the evaluation session, the participants were randomly asked to hold a total of 42 gestures. The requested gestures were selected at random every 5 seconds (the time series associated with an evaluation session thus lasted 210 seconds). Each participant recorded a minimum of 6 evaluation sessions (twice per recording session). Importantly, for the evaluation session, the transition between each gesture was recorded, yielding a continuous time series. 

For this work, two datasets were derived from the Long-Term 3DC Dataset: The EMG Artificial Dataset and the EMG Dataset. 

The EMG Artificial Dataset was created from the original training sessions by concatenating these gesture recordings together to form a continuous time series for each participant. The artificial training set consists of training sessions recorded from ten participants. The validation set consists of 15 time series made by gestures from four participants, while the test sets includes 30 time series from eight participants. 

The EMG Dataset was created from the evaluation sessions. The evaluation runs of six participants is used as the training set, the evaluation runs of five participants is used as the validation set, and the evaluation runs for the remaining nine participants is used as the test set. This dataset will be referred to as the EMG Dataset.



\subsection{Hyperparameter Selection}
The segmentation algorithms are divided into two categories: online and offline.

Even if they are not fully online, LS-USS, FLUSS and LFMD are also included in the online category as they can be made to work $\varepsilon$ real-time when using a temporal constraint.
The hyperparameter optimization is done using grid search by leveraging the previously described training and validation sets for the three considered datasets. 
The step size, used in LFMD, decides the amount of overlap between subsequences. 
For the segmentation algorithms that uses an autoencoder (LS-USS and LFMD), the fully connected and convolutional model is included in the hyperparameter search. 
For the offline algorithms the extraction algorithms REA and LREA are included in the hyperparameter search.
Four data scaling techniques are applied to data for comparing the segmentation algorithms on the different datasets. These methods use the implementation provided in scikit-learn \cite{noauthor_scikit-learn_nodate} and are widely used for machine learning and data mining tasks. The scaling is done independently over all channels, and the statistics used for scaling the data are computed using only the training data. 
The choice of scaling method is also incorporated into the hyperparameter search.

The mean distance between the change-points from the training set is calculated for use as the local window size for scaling the CAC when using the LREA and LTEA extractors. The training set is also utilized for training the autoencoders. As the autoencoders require both training and validation sets, 80\% of the training set is used as training examples, while the remaining 20\% is used to validate the autoencoder. The validation portion of the dataset is used to find the optimal hyperparameters of the segmentation algorithms. After the best performing model configurations on the validation set are found, the test set is used to make the final comparisons between the different segmentation algorithms. 

For a detailed presentation of the considered hyperparameters see Table~\ref{table:offline_hyperparameters} and Table~\ref{table:online_hyperparameters} in the Appendix.
Furthermore, Table~\ref{table:offline_hyperparameters_selected} and Table~\ref{table:online_hyperparameters_selected} in the appendix shows the hyperparameters selected for the offline and online evaluation respectively. 

\subsection{Evaluation metrics}
The performance of the offline algorithms are evaluated based on the ScoreRegimes from Table 3 in \cite{FLUSS}. The definition of the ScoreRegimes is as follows:
\begin{equation}
\label{eq:regime_score}
\text { ScoreRegimes }=\frac{\sum_{i=1}^{N_{G T}}\left|C P_{p r e d}-C P_{a c t u a l}\right|}{N_{G T} * n}
\end{equation}
Where $N_{G T}$ is the number of ground truth change-points, and $n$ is the length of the time series. 

Note that the ScoreRegimes requires knowing the number of ground truth change-points and are thus ill-adapted to evaluate online CPD algorithms as they might identify a different number of change-points compared to the ground truth. Thus, the online algorithms are evaluated using the Prediction Loss mean absolute error (MAE), which is a slight modification of equation 12 in \cite{LFMD}. The Prediction Loss MAE used in this work is defined as follows:

\begin{equation}
\label{eq:prediction_loss}
\text { Prediction loss } M A E=\left|1-\frac{N_{\text {pred }}}{N_{G T}}\right| \times M A E
\end{equation}
Where $N_{\text {pred }}$ is the number of predicted change-points and $N_{G T}$ is the number of ground truth change-points. Thus, instead of relying on pre-defined change-points, the Prediction loss MAE weights the mean absolute error with the prediction ratio. Importantly, as the metrics used for the offline and online case are not related, it is also not possible to make comparisons between the online and offline algorithms' performance. 





As suggested in~\cite{statistical_test_comparison}, a two-step statistical procedure is applied to compare LS-USS against the relevant CPD algorithms. First, Friedman's test ranks the algorithms amongst each other. Then, Holm's post-hoc test is applied using the best ranked method as a comparison basis. The null hypothesis of the post hoc test is that the performance of the two models is the same. The null hypothesis is rejected when p$<$0.05.


\section{Results}
\label{sec:results}
The comparisons in this section are made on three datasets in both the offline and online setting: UCI, EMG artificial, and EMG. 
\subsection{Offline}
In this subsection, for each dataset and participant, the full time series was available when performing unsupervised segmentation and the number of change-points in each time series was also known. Table~\ref{table:results_offline} presents the results obtained in the offline setting. 

\begin{table}[]
\begin{center}
\begin{tabular}{@{}cccc@{}}
\toprule
\multicolumn{4}{c}{{\ul \textbf{UCI}}} \\ \midrule
 & \textbf{LS-USS} & \textbf{FLUSS} & \textbf{LFMD} \\
Mean & \textbf{0.00687} & 0.00931 & 0.01580 \\
Friedman Rank & \textbf{1.33} & 2.07 & 2.60 \\
H0 (Adjusted p-value) & \textbf{-} & 0 (0.04461) & 0 (0.00105) \\ \midrule
\multicolumn{4}{c}{{\ul \textbf{EMG Artificial}}} \\ \midrule
 & \textbf{LS-USS} & \textbf{FLUSS} & \textbf{LFMD} \\
Mean & \textbf{0.00214} & 0.00718 & 0.01839 \\
Friedman Rank & \textbf{1.03} & 1.97 & 3.00 \\
H0 (Adjusted p-value) & \textbf{-} & 0 (0.00030) & 0 (\textless{}0.00001) \\ \midrule
\multicolumn{4}{c}{{\ul \textbf{EMG}}} \\ \midrule
 & \textbf{LS-USS} & \textbf{FLUSS} & \textbf{LFMD} \\
Mean & \textbf{0.00452} & 0.00593 & 0.00715 \\
Friedman Rank & \textbf{1.00} & 2.00 & 3.50 \\
H0 (Adjusted p-value) & \textbf{-} & 0 (0.01431) & 0 (\textless{}0.00001) \\ \bottomrule
\end{tabular}
\label{table:results_offline}
\caption{Comparison between the considered CPD algorithms in the offline setting.}
\end{center}
\end{table}

\subsection{Online}
In this subsection, data was fed to the CPD algorithms as if they were acquired in real-time. Further, the total number of change-points in a time series was not given. Table~\ref{table:results_online} presents the results obtained in the online setting. 

\begin{table*}[]
\begin{center}
\begin{tabular}{cccccc}
\hline
\multicolumn{6}{c}{{\ul \textbf{UCI}}} \\ \hline
 & \textbf{LS-USS} & \textbf{LS-USS Online} & \textbf{FLUSS} & \textbf{FLOSS} & \textbf{LFMD} \\
Mean & \textbf{45.54} & 114.82 & 47.27 & 87.36 & 51.99 \\
Friedman Rank & \textbf{2.50} & 4.20 & 2.56 & 3.12 & 2.60 \\
H0 (Adjusted p-value) & \textbf{-} & 0 (0.01294) & 1 & 1 & 1 \\ \hline
\multicolumn{6}{c}{{\ul \textbf{EMG Artificial}}} \\ \hline
 & \textbf{LS-USS} & \textbf{LS-USS Online} & \textbf{FLUSS} & \textbf{FLOSS} & \textbf{LFMD} \\
Mean & \textbf{6.64} & 16.10 & 26.73 & 86.44 & 97.54 \\
Friedman Rank & \textbf{1.82} & 2.28 & 2.80 & 3.88 & 4.22 \\
H0 (Adjusted p-value) & \textbf{-} & 1 & 0 (0.03202) & 0 (\textless{}0.00001) & 0 (\textless{}0.00001) \\ \hline
\multicolumn{6}{c}{{\ul \textbf{EMG}}} \\ \hline
 & \textbf{LS-USS} & \textbf{LS-USS Online} & \textbf{FLUSS} & \textbf{FLOSS} & \textbf{LFMD} \\
Mean & \textbf{121.97} & 164.68 & 251.41 & 163.71 & 388.82 \\
Friedman Rank & \textbf{1.50} & 2.50 & 3.70 & 2.55 & 4.75 \\
H0 (Adjusted p-value) & \textbf{-} & 1 & 0 (0.00003) & 1 & 0 (\textless{}0.00001) \\ \hline
\end{tabular}
\label{table:results_online}
\caption{Comparison between the considered CPD algorithms in the online setting.}
\end{center}
\end{table*}

\section{Discussion}
\label{sec:discussion}

\subsubsection{Observations from the hyperparameter search}
Table~\ref{table:offline_hyperparameters_selected} in the appendix shows that LREA is chosen seven out of nine times for the three algorithms that are being compared on the three different datasets. This indicates that scaling the CAC based on the local statistics is useful when dealing with longer time series data. Another interesting finding from the hyperparameter selection is that the autoencoder selected by the hyperparameter search for the LS-USS models (both offline and online) varies between the datasets; some use the less complex, fully connected model, while others use the convolutional model. As alluded to in~\cite{LFMD}, using a smaller model with only two hidden layers might be beneficial for feature representation as it could lead to a more general way to represent the data. The drawback of simpler autoencoder models is that one risks losing some of the information needed for segment differentiation. Note that an in-depth analysis of the impact of the autoencoder's architecture was outside the scope of the current work and will be the focus of future works. 

\subsubsection{Offline}
As shown in Table~\ref{table:results_offline}, in the offline setting LS-USS is shown to systematically and significantly outperform the other segmentation algorithms for all three considered datasets. This indicates that there is a clear advantage in terms of overall performance of the CDP algorithm in learning a representation which can consider the multidimensional representation of the data simultaneously compared to using FLUSS directly. Further, the fact that LFMD was the worst performing CDP algorithm illustrates the usefulness of leveraging the latent space matrix profile to segment the time serie. 

\subsubsection{Online}
The highest-ranked algorithm on the UCI dataset is LS-USS, but the difference with the other algorithms is not significant. $\varepsilon$-real time LS-USS is also the best-ranked algorithm on both the EMG and EMG Artificial datasets and the difference in performance between it and the other algorithms is also significant (except for LS-USS online on the EMG Artificial dataset and LS-USS online and FLOSS on the EMG dataset). Overall, the $\varepsilon$-real time LS-USS LTEA model seems to be the most consistent performing online segmentation algorithm across all the datasets.

\subsection{Limitations}
The segmentation outcomes for each of the algorithms are dependent on the hyperparameter search. Therefore, it would be beneficial to do a more in-depth hyperparameter search that includes sampling the hyperparameters from predefined distributions instead of a simple grid search. Further, a more in-depth evaluation of the hyperparameter selection sensitivity will be conducted in future works.

An additional limitation of LS-USS compared to FLUSS is that despite the improvement observed in terms of performance, it does come at the cost of an increase in the hyperparameters due to the use of an autoencoder. 

The experiments conducted in this work have mainly been on multidimensional data where the channels are from similar sensors, similar sampling rates, and numerical data only. Therefore, how the algorithm performs when considering time series data containing channels from a broader range of sources with both numerical and categorical variables remains to be evaluated. Furthermore, doing an in-depth analysis on how the number of channels affects the performance of LS-USS will also be considered in future works. 

\section{Conclusion}
\label{sec:conclusion}
The main contribution of this work is the segmentation algorithms LS-USS and LS-USS online. Through extensive testing conducted on both artificial and real-world datasets from various domains and sensors, it was found that LS-USS generally delivers on par or better segmentation scores compared to other state-of-the-art algorithms such as FLUSS/FLOSS and LFMD. The LS-USS algorithms have many desirable properties. They can be implemented both online and in an “anytime” fashion. They are domain agnostic (beyond knowing the order of time scale considered for change-points) and do not need extensive tuning of hyperparameters to achieve state-of-the-art performance. Further, they do not make any statistical assumptions about the input data. The LSMP component used in LS-USS also shows that it is possible to calculate a temporarily constrained matrix profile from feature vectors by exploiting highly parallelized hardware. The same methods used for constructing the LSMP can be used on any vector-based Representation Learning algorithms~\cite{representation_learning}, which represents an excellent potential direction for future research. 

The extraction algorithms LREA and LTEA presented in this work also show potential. LREA usually outperformed the more global REA algorithm, which shows that scaling the CAC based on the local statistics is highly useful, especially for long time series data. The online extraction algorithm LTEA uses the same CAC scaling as LREA but uses a threshold to identify the change-points instead of extracting the $n$ lowest “valley” points. Consequently, in contrast to REA and LREA, LTEA does not need any information on the number of change-points to extract, making it applicable to a wider array of real-world segmentation problems.

\bibliographystyle{IEEEtran}
\bibliography{main}

\appendices

\section{Hyperparameters Search and Selection}

Table \ref{table:offline_hyperparameters} and Table \ref{table:online_hyperparameters} shows the hyperparameters considered for the offline algorithm and online algorithms respectively. 

\begin{table*}[!htpb]
\centering
\begin{tabular}{cccccccccc}
\hline
\textbf{Offline Models} & \textbf{Data Scalers} & \multicolumn{2}{c}{\textbf{NW}} & \multicolumn{2}{c}{\textbf{TC}} & \multicolumn{2}{c}{\textbf{Step-Size}} & \textbf{Autoencoder} & \textbf{Extraction Algorithm} \\ \hline
 & All Datasets & \multicolumn{2}{c}{All datasets} \  & \begin{tabular}[c]{@{}c@{}}EMG \\ Datasets\end{tabular} & UCI & \multicolumn{2}{c}{All datasets} & All Datasets & All Datasets \\
LS-USS & \begin{tabular}[c]{@{}c@{}}No Scaler\\ Standard Scaler\\ Robust Scaler\\ Min Max Scaler\end{tabular} & \begin{tabular}[c]{@{}c@{}}50\\ 100\\ 150\\ 200\end{tabular} & \begin{tabular}[c]{@{}c@{}}300\\ 400\\ 500\end{tabular} & \begin{tabular}[c]{@{}c@{}}1000\\ 1500\\ 2000\end{tabular} & \begin{tabular}[c]{@{}c@{}}800\\ 1200\\ 1600\end{tabular} & - & - & \begin{tabular}[c]{@{}c@{}}Fully Connected\\ Convolutional\end{tabular} & \begin{tabular}[c]{@{}c@{}}REA\\ LREA\end{tabular} \\ \hline
FLUSS & \begin{tabular}[c]{@{}c@{}}No Scaler\\ Standard Scaler\\ Robust Scaler\\ Min Max Scaler\end{tabular} & \begin{tabular}[c]{@{}c@{}}50\\ 100\\ 150\\ 200\end{tabular} & \begin{tabular}[c]{@{}c@{}}300\\ 400\\ 500\end{tabular} & \begin{tabular}[c]{@{}c@{}}1000\\ 1500\\ 2000\end{tabular} & \begin{tabular}[c]{@{}c@{}}800\\ 1200\\ 1600\end{tabular} & - & - & - & \begin{tabular}[c]{@{}c@{}}REA\\ LREA\end{tabular} \\ \hline
LFMD & \begin{tabular}[c]{@{}c@{}}No Scaler\\ Standard Scaler\\ Robust Scaler\\ Min Max Scaler\end{tabular} & \begin{tabular}[c]{@{}c@{}}50\\ 100\\ 150\\ 200\end{tabular} & \begin{tabular}[c]{@{}c@{}}300\\ 400\\ 500\end{tabular} & \begin{tabular}[c]{@{}c@{}}1000\\ 1500\\ 2000\end{tabular} & \begin{tabular}[c]{@{}c@{}}800\\ 1200\\ 1600\end{tabular} & \begin{tabular}[c]{@{}c@{}}25\\ 50\\ 100\\ 150\end{tabular} & \begin{tabular}[c]{@{}c@{}}200\\ 250\\ 350\\ 500\end{tabular} & \begin{tabular}[c]{@{}c@{}}Fully Connected\\ Convolutional\end{tabular} & \begin{tabular}[c]{@{}c@{}}REA\\ LREA\end{tabular} \\ \hline
\end{tabular}
\caption{Hyperparameters considered for the CPD algorithms in the offline setting. NW is the sub-sequence length. TC is the temporal constraint.}
\label{table:offline_hyperparameters}
\end{table*}

\begin{table*}[]
\centering
\begin{tabular}{@{}cccccccccccc@{}}
\toprule
\textbf{Online Models} & \textbf{Data Scalers} & \multicolumn{2}{c}{\textbf{NW}} & \multicolumn{2}{c}{\textbf{TC}} & \multicolumn{2}{c}{\textbf{Step-size}} & \textbf{Autoencoder} & \multicolumn{2}{c}{\textbf{NW}} & \textbf{\begin{tabular}[c]{@{}c@{}}Extraction\\ Algorithm\end{tabular}} \\ \midrule
 & All Datasets & \multicolumn{2}{c}{All datasets} & \begin{tabular}[c]{@{}c@{}}EMG\\  datasets\end{tabular} & UCI & \multicolumn{2}{c}{All datasets} & All Datasets & \multicolumn{2}{c}{All datasets} & All Datasets \\
\begin{tabular}[c]{@{}c@{}}LS-USS \\ ($\varepsilon$-real time)\end{tabular} & \begin{tabular}[c]{@{}c@{}}No Scaler\\ Standard Scaler\\ Robust Scaler\\ Min Max Scaler\end{tabular} & \begin{tabular}[c]{@{}c@{}}50\\ 100\\ 150\\ 200\end{tabular} & \begin{tabular}[c]{@{}c@{}}300\\ 400\\ 500\end{tabular} & \begin{tabular}[c]{@{}c@{}}1000\\ 1500\\ 2000\end{tabular} & \begin{tabular}[c]{@{}c@{}}800\\ 1200\\ 1600\end{tabular} & - & - & \begin{tabular}[c]{@{}c@{}}Fully Connected\\ Convolutional\end{tabular} & \begin{tabular}[c]{@{}c@{}}-0.5\\ -1.0\\ -1.5\end{tabular} & \begin{tabular}[c]{@{}c@{}}-2.0\\ -2.5\\ -3.0\end{tabular} & LTEA \\ \midrule
LS-USS - Online & \begin{tabular}[c]{@{}c@{}}No Scaler\\ Standard Scaler\\ Robust Scaler\\ Min Max Scaler\end{tabular} & \begin{tabular}[c]{@{}c@{}}50\\ 100\\ 150\\ 200\end{tabular} & \begin{tabular}[c]{@{}c@{}}300\\ 400\\ 500\end{tabular} & \begin{tabular}[c]{@{}c@{}}1000\\ 1500\\ 2000\end{tabular} & \begin{tabular}[c]{@{}c@{}}800\\ 1200\\ 1600\end{tabular} & - & - & \begin{tabular}[c]{@{}c@{}}Fully Connected\\ Convolutional\end{tabular} & \begin{tabular}[c]{@{}c@{}}-0.5\\ -1.0\\ -1.5\end{tabular} & \begin{tabular}[c]{@{}c@{}}-2.0\\ -2.5\\ -3.0\end{tabular} & LTEA \\ \midrule
\begin{tabular}[c]{@{}c@{}}FLUSS \\ ($\varepsilon$-real time)\end{tabular} & \begin{tabular}[c]{@{}c@{}}No Scaler\\ Standard Scaler\\ Robust Scaler\\ Min Max Scaler\end{tabular} & \begin{tabular}[c]{@{}c@{}}50\\ 100\\ 150\\ 200\end{tabular} & \begin{tabular}[c]{@{}c@{}}300\\ 400\\ 500\end{tabular} & \begin{tabular}[c]{@{}c@{}}1000\\ 1500\\ 2000\end{tabular} & \begin{tabular}[c]{@{}c@{}}800\\ 1200\\ 1600\end{tabular} & - & - & - & \begin{tabular}[c]{@{}c@{}}-0.5\\ -1.0\\ -1.5\end{tabular} & \begin{tabular}[c]{@{}c@{}}-2.0\\ -2.5\\ -3.0\end{tabular} & LTEA \\ \midrule
FLOSS & \begin{tabular}[c]{@{}c@{}}No Scaler\\ Standard Scaler\\ Robust Scaler\\ Min Max Scaler\end{tabular} & \begin{tabular}[c]{@{}c@{}}50\\ 100\\ 150\\ 200\end{tabular} & \begin{tabular}[c]{@{}c@{}}300\\ 400\\ 500\end{tabular} & \begin{tabular}[c]{@{}c@{}}1000\\ 1500\\ 2000\end{tabular} & \begin{tabular}[c]{@{}c@{}}800\\ 1200\\ 1600\end{tabular} & - & - & - & \begin{tabular}[c]{@{}c@{}}-0.5\\ -1.0\\ -1.5\end{tabular} & \begin{tabular}[c]{@{}c@{}}-2.0\\ -2.5\\ -3.0\end{tabular} & LTEA \\ \midrule
LFMD & \begin{tabular}[c]{@{}c@{}}No Scaler\\ Standard Scaler\\ Robust Scaler\\ Min Max Scaler\end{tabular} & \begin{tabular}[c]{@{}c@{}}50\\ 100\\ 150\\ 200\end{tabular} & \begin{tabular}[c]{@{}c@{}}300\\ 400\\ 500\end{tabular} & \begin{tabular}[c]{@{}c@{}}1000\\ 1500\\ 2000\end{tabular} & \begin{tabular}[c]{@{}c@{}}800\\ 1200\\ 1600\end{tabular} & \begin{tabular}[c]{@{}c@{}}25\\ 50\\ 100\\ 150\end{tabular} & \begin{tabular}[c]{@{}c@{}}200\\ 250\\ 350\\ 500\end{tabular} & \begin{tabular}[c]{@{}c@{}}Fully Connected\\ Convolutional\end{tabular} & \begin{tabular}[c]{@{}c@{}}-0.5\\ -1.0\\ -1.5\end{tabular} & \begin{tabular}[c]{@{}c@{}}-2.0\\ -2.5\\ -3.0\end{tabular} & LTEA \\ \bottomrule
\end{tabular}
\caption{Hyperparameters considered for the CPD algorithms in the online setting. NW is the sub-sequence length. TC is the temporal constraint. The threshold parameter is the threshold used in the LTEA extraction algorithm.}
\label{table:online_hyperparameters}
\end{table*}
\clearpage
Tables \ref{table:offline_hyperparameters_selected} and \ref{table:online_hyperparameters_selected} shows the hyperparameters selected for the offline and online algorithms after doing hyperparameter search. The selected parameters is the ones used for the comparisons between the different models in Section~\ref{sec:results} and \ref{sec:discussion}.

\begin{table*}[!hpb]
\begin{center}
\begin{tabular}{@{}ccccccc@{}}
\toprule
\multicolumn{7}{c}{{\ul \textbf{UCI}}} \\ \midrule
\textbf{CPD Algorithm} & \textbf{Extraction Algorithm} & \textbf{Data Scaler} & \textbf{NW} & \textbf{TC} & \textbf{Step-size} & \textbf{Autoencoder} \\
LFMD & LREA & Robust Scaler & 50 & - & 250 & Fully Connected \\
LS-USS & REA & Min Max Scaler & 50 & 800 & 1 & Fully Connected \\
FLUSS & LREA & Robust Scaler & 50 & 1200 & 1 & - \\ \midrule
\multicolumn{7}{c}{{\ul \textbf{EMG Artificial}}} \\ \midrule
\textbf{CPD Algorithm} & \textbf{Extraction Algorithm} & \textbf{Data Scaler} & \textbf{NW} & \textbf{TC} & \textbf{Step-size} & \textbf{Autoencoder} \\
LFMD & REA & Standard Scaler & 200 & - & 500 & Conv. Model \\
LS-USS & LREA & Standard Scaler & 50 & 1500 & 1 & Conv. Model \\
FLUSS & LREA & Standard Scaler & 50 & 1500 & 1 & - \\ \midrule
\multicolumn{7}{c}{{\ul \textbf{EMG}}} \\ \midrule
\textbf{CPD Algorithm} & \textbf{Extraction Algorithm} & \textbf{Data Scaler} & \textbf{NW} & \textbf{TC} & \textbf{Step-size} & \textbf{Autoencoder} \\
LFMD & LREA & Standard Scaler & 50 & - & 100 & Fully Connected \\
LS-USS & LREA & Robust Scaler & 50 & 2000 & 1 & Conv. Model \\
FLUSS & LREA & Standard Scaler & 50 & 1500 & 1 & - \\ \bottomrule
\end{tabular}
\end{center}
\caption{Hyperparameters selected for the compared CPD algorithms in the offline setting. NW is the sub-sequence length. TC is the temporal constraint.}
\label{table:offline_hyperparameters_selected}
\end{table*}

\begin{table*}[!htpb]
\begin{center}
\begin{tabular}{@{}cccccccc@{}}
\toprule
\multicolumn{8}{c}{{\ul \textbf{UCI}}} \\ \midrule
\textbf{CPD Algorithm} & \textbf{Extraction Algorithm} & \textbf{Data Scaler} & \textbf{NW} & \textbf{TC} & \textbf{Step-size} & \textbf{Autoencoder} & \textbf{Threshold} \\
LFMD & LTEA & Min Max Scaler & 100 & - & 100 & Fully Connected & -2 \\
FLUSS & LTEA & Robust Scaler & 50 & 1600 & 1 & - & -1 \\
FLOSS & LTEA & Robust Scaler & 150 & 1600 & 1 & - & -1 \\
LS-USS & LTEA & Min Max Scaler & 150 & 800 & 1 & Fully Connected & -1.5 \\
LS-USS Online & LTEA & Robust Scaler & 50 & 1200 & 1 & Fully Connected & -0.5 \\ \midrule
\multicolumn{8}{c}{{\ul \textbf{EMG Artificial}}} \\ \midrule
\textbf{CPD Algorithm} & \textbf{Extraction Algorithm} & \textbf{Data Scaler} & \textbf{NW} & \textbf{TC} & \textbf{Step-size} & \textbf{Autoencoder} & \textbf{Threshold} \\
LFMD & LTEA & Standard Scaler & 300 & - & 100 & Fully Connected & -0.5 \\
FLUSS & LTEA & Standard Scaler & 50 & 1500 & 1 & - & -1 \\
FLOSS & LTEA & Standard Scaler & 100 & 2000 & 1 & - & -0.5 \\
LS-USS & LTEA & Standard Scaler & 50 & 2000 & 1 & Conv. Model & -0.5 \\
LS-USS Online & LTEA & Standard Scaler & 50 & 1500 & 1 & Conv. Model & -1 \\ \midrule
\multicolumn{8}{c}{{\ul \textbf{EMG}}} \\ \midrule
\textbf{CPD Algorithm} & \textbf{Extraction Algorithm} & \textbf{Data Scaler} & \textbf{NW} & \textbf{TC} & \textbf{Step-size} & \textbf{Autoencoder} & \textbf{Threshold} \\
LFMD & LTEA & Standard Scaler & 500 & - & 500 & Fully Connected & -1.5 \\
FLUSS & LTEA & Standard Scaler & 50 & 1000 & 1 & - & -2 \\
FLOSS & LTEA & Standard Scaler & 200 & 1000 & 1 & - & -2 \\
LS-USS & LTEA & Standard Scaler & 300 & 1000 & 1 & Fully Connected & -2 \\
LS-USS Online & LTEA & Standard Scaler & 200 & 1500 & 1 & Fully Connected & -2 \\ \bottomrule
\end{tabular}
\end{center}
\caption{Hyperparameters selected for the compared CPD algorithms in the online setting. NW is the sub-sequence length. TC is the temporal constraint. The threshold parameter is the threshold used in the LTEA extraction algorithm.}
\label{table:online_hyperparameters_selected}
\end{table*}

\end{document}